\documentclass[11pt]{article}

\usepackage[final]{acl}

\usepackage{times}
\usepackage{latexsym}

\usepackage[T1]{fontenc}
\usepackage[utf8]{inputenc}

\usepackage{microtype}

\usepackage{inconsolata}

\usepackage{graphicx}

\usepackage{algorithm}
\usepackage{amssymb}
\usepackage{amsmath}
\usepackage{algorithmic}
\usepackage{enumitem}
\usepackage{booktabs}
\usepackage{siunitx} 
\usepackage{multirow}
\usepackage{subcaption}
\usepackage{bm}
\usepackage{url}
\usepackage{listings}
\usepackage{tcolorbox}
\tcbuselibrary{listings}

\usepackage{colortbl}
\definecolor{lightgray}{gray}{0.9}

\lstdefinestyle{mypromptstyle}{
    backgroundcolor=\color{black!5},   
    commentstyle=\color{green!50!black}, 
    keywordstyle=\color{blue},         
    numberstyle=\tiny\color{gray},     
    stringstyle=\color{purple},        
    basicstyle=\ttfamily\footnotesize, 
    breakatwhitespace=false,
    breaklines=true,
    captionpos=b,
    keepspaces=true,
    numbers=none,                      
    numbersep=5pt,
    showspaces=false,
    showstringspaces=false,
    showtabs=false,
    tabsize=2,
    frame=single,
    framextopmargin=5pt,               
    framexbottommargin=5pt,            
    framexleftmargin=0pt,              
    framexrightmargin=0pt 
}

\definecolor{mybrown}{HTML}{8B4513}
\definecolor{mycream}{HTML}{FFF8E7}

\definecolor{codebg}{RGB}{250,250,250}       
\definecolor{codekw}{RGB}{33, 74, 135}        
\definecolor{codecomment}{RGB}{63, 127, 95}    
\definecolor{codestring}{RGB}{163, 21, 21}     
\definecolor{codeid}{RGB}{0,0,0}  

\lstdefinestyle{mypython}{
    language=Python,
    backgroundcolor=\color{codebg},
    basicstyle=\ttfamily\footnotesize,
    keywordstyle=\color{codekw}\bfseries,
    commentstyle=\color{codecomment}\itshape,
    stringstyle=\color{codestring},
    identifierstyle=\color{codeid},
    showstringspaces=false,
    breaklines=true,
    tabsize=4,
    numbers=none,
    numberstyle=\tiny\color{gray},
    stepnumber=1,
    numbersep=8pt,
    frame=single,
    rulecolor=\color{gray},
    frameround=tttt
}

\newtcblisting{codebox}[2][]{
    listing only,
    listing options={style=mypython},
    enhanced,
    colback=mycream,
    colframe=mybrown,
    fonttitle=\bfseries,
    colbacktitle=mybrown,
    coltitle=white,
    title=#2,
    boxed title style={arc=0mm, outer arc=1mm,width=\linewidth},
    arc=0mm,
    left=0mm,
    #1
}

\title{Solver-Independent Automated Problem Formulation via LLMs for High-Cost Simulation-Driven Design}

\author{
Yuchen Li\textsuperscript{\rm 1}, Handing Wang\textsuperscript{\rm 1}, Bing Xue\textsuperscript{\rm 2}, Mengjie Zhang\textsuperscript{\rm 2} \and Yaochu Jin\textsuperscript{\rm 3} \\
\textsuperscript{\rm 1}School of Artificial Intelligence, Xidian University, China \\
\textsuperscript{\rm 2}School of Engineering and Computer Science, Victoria University of Wellington, New Zealand \\
\textsuperscript{\rm 3}School of Engineering, Westlake University, China \\
\texttt{ycli\_7@stu.xidian.edu.cn, hdwang@xidian.edu.cn, bing.xue@ecs.vuw.ac.nz}\\
\texttt{mengjie.zhang@ecs.vuw.ac.nz, jinyaochu@westlake.edu.cn}\\
}

\begin{document}
\maketitle

\begin{abstract}
In the high-cost simulation-driven design domain, translating ambiguous design requirements into a mathematical optimization formulation is a bottleneck for optimizing product performance. 
This process is time-consuming and heavily reliant on expert knowledge. 
While large language models (LLMs) offer potential for automating this task, existing approaches either suffer from poor formalization that fails to accurately align with the design intent or rely on solver feedback for data filtering, which is unavailable due to the high simulation costs.
To address this challenge, we propose automated problem formulation (APF), a solver-independent framework that utilizes LLMs to convert engineers' natural language requirements into executable optimization models.
The core of this framework is an innovative pipeline for automatically generating high-quality data, which overcomes the difficulty of constructing suitable fine-tuning datasets in the absence of high-cost solver feedback with the help of data generation and test instance annotation. 
The generated high-quality dataset is used to perform supervised fine-tuning on LLMs, significantly enhancing their ability to generate accurate and executable optimization problem formulations. 
Experimental results on antenna design demonstrate that APF significantly outperforms the existing methods in both the accuracy of requirement formalization and the quality of resulting radiation efficiency curves in meeting the design goals.
\end{abstract}

\section{Introduction}

\begin{figure}[t]
    \centering
    \includegraphics[width=0.99\linewidth]{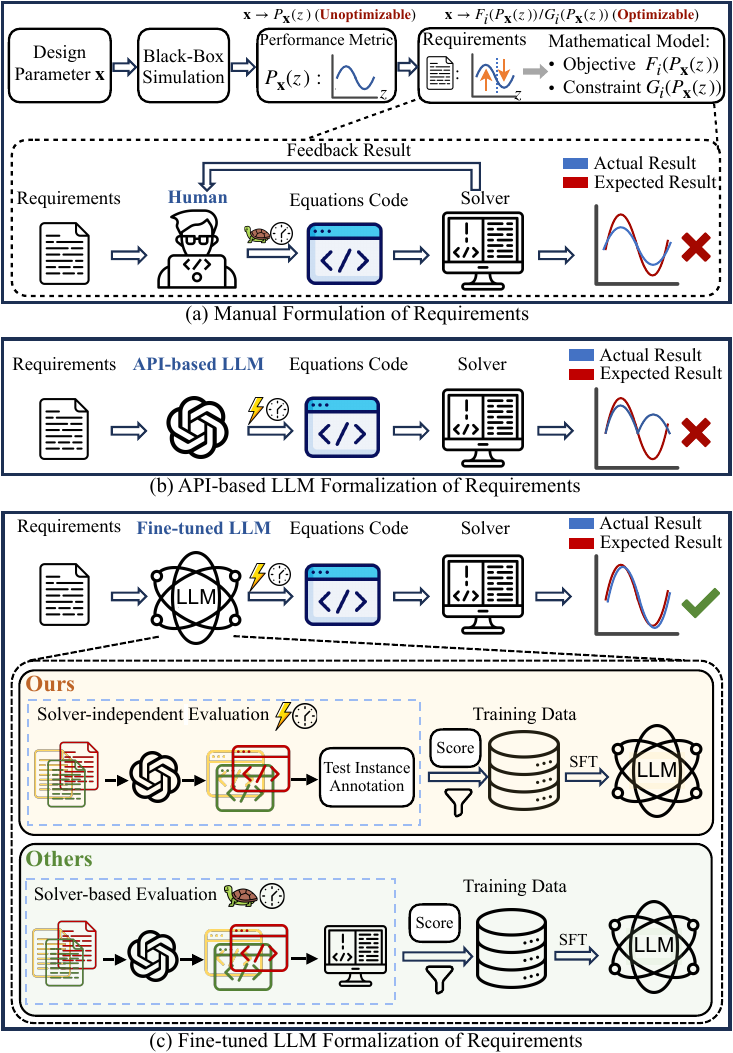}

    \caption{Formalizing Requirements in High-Cost Simulation-Driven Design: From Manual Expertise to LLM-based Workflow}
    \label{fig:work_diff}
\end{figure}

High-cost simulation-driven design is prevalent in numerous fields, such as antenna \cite{Kouhalvandi25}, aerospace \cite{wissink2025multi,liMachineLearningAerodynamic2022}, microelectronics \cite{NEURIPS2024_fb23cf87}, and robotics \cite{tevet2025closd}.
As illustrated in Figure \ref{fig:work_diff}(a), a common task in product design across these fields is to optimize design parameters, ensuring that the performance distribution under given evaluation variables (e.g., frequency, angle) satisfies specific design requirements. 
In practice, this performance distribution is typically obtained through high-fidelity simulations and manifests itself as high-dimensional curves, which are difficult for optimization algorithms to use directly. 
Therefore, it is necessary to formalize the design requirements into an executable mathematical model to serve as the objectives or constraints for an optimization algorithm. 
However, since design requirements are often provided in unstructured natural language, and the formalization process is time-consuming, and highly dependent on engineering expertise, automating this procedure remains a significant challenge.

The development of large language models (LLMs), such as GPT \cite{openaiGPT4TechnicalReport2024}, Gemini \cite{teamGeminiFamilyHighly2025, teamGemini15Unlocking2024}, and DeepSeek \cite{deepseek-aiDeepSeekV3TechnicalReport2025}, offers a promising avenue to automatically formulate optimization problems. 
The existing work in this domain can be broadly divided into prompt-based (Figure~\ref{fig:work_diff}(b)) and fine-tuning-based (Figure~\ref{fig:work_diff}(c)) methods.
Early work such as Chain-of-Experts \cite{xiaoChainofExpertsWhenLLMs2023} and OptiMUS \cite{ahmaditeshniziOptiMUSScalableOptimization2024} explore prompt-based methods, where the model is guided to generate optimization problems by carefully designing input prompts. 
Fine-tuning-based methods such as LLaMoCo \cite{maLLaMoCoInstructionTuning2024}, ORLM \cite{huangORLMCustomizableFramework2025}, LLMOPT \cite{jiangLLMOPTLearningDefine2024}, and SIRL \cite{chenSolverInformedRLGrounding2025} improve the performance of LLMs by training on task-specific data, enabling better understanding and generation of optimization problems.
These studies show that fine-tuned models with moderate sizes (e.g., 7 billion parameters) can sometimes perform better than larger general-purpose models like GPT-4. 
This highlights the advantage of fine-tuning in improving the accuracy and reliability of problem formulation for optimization tasks.

Although previous studies have explored methods for automatically converting optimization problems described in natural language into mathematical models using LLMs, most of these methods focus on operational optimization problems such as linear programming and integer programming, which differ significantly in their problem description and evaluation costs from the high-cost simulation-driven design scenarios.
Therefore, the application of these methods is limited in complex industrial design scenarios.
In particular, prompt-based methods struggle to accurately identify objectives and constraints when faced with natural language requirements that are vague or heavily reliant on domain-specific knowledge. 
While fine-tuning-based methods improve the LLM's ability to handle structured optimization tasks, they face limitations in high-cost simulation-driven design scenarios where the cost of solver feedback prevents effective data quality filtering.

To address the modeling challenges in high-cost simulation-driven design, we propose a framework for automated problem formulation, called APF. 
This framework utilizes a fine-tuned LLM to automatically convert engineers' natural language requirements into accurate and executable optimization problem formulations.
Our method comprises three main contributions:
(1) We develop an automated framework combining data augmentation and test instance annotation to construct high-quality datasets for LLMs fine-tuning.
(2) We introduce a solver-independent evaluation module that utilizes LLMs reasoning to assess formulation quality, eliminating the need for expensive solver feedback.
(3) We apply SFT on the dataset and evaluate APF on an industrial optimization task. Results demonstrate that our method consistently outperforms existing approaches in accuracy.

\begin{figure*}[t] 
    \centering
    \includegraphics[width=0.99\linewidth]{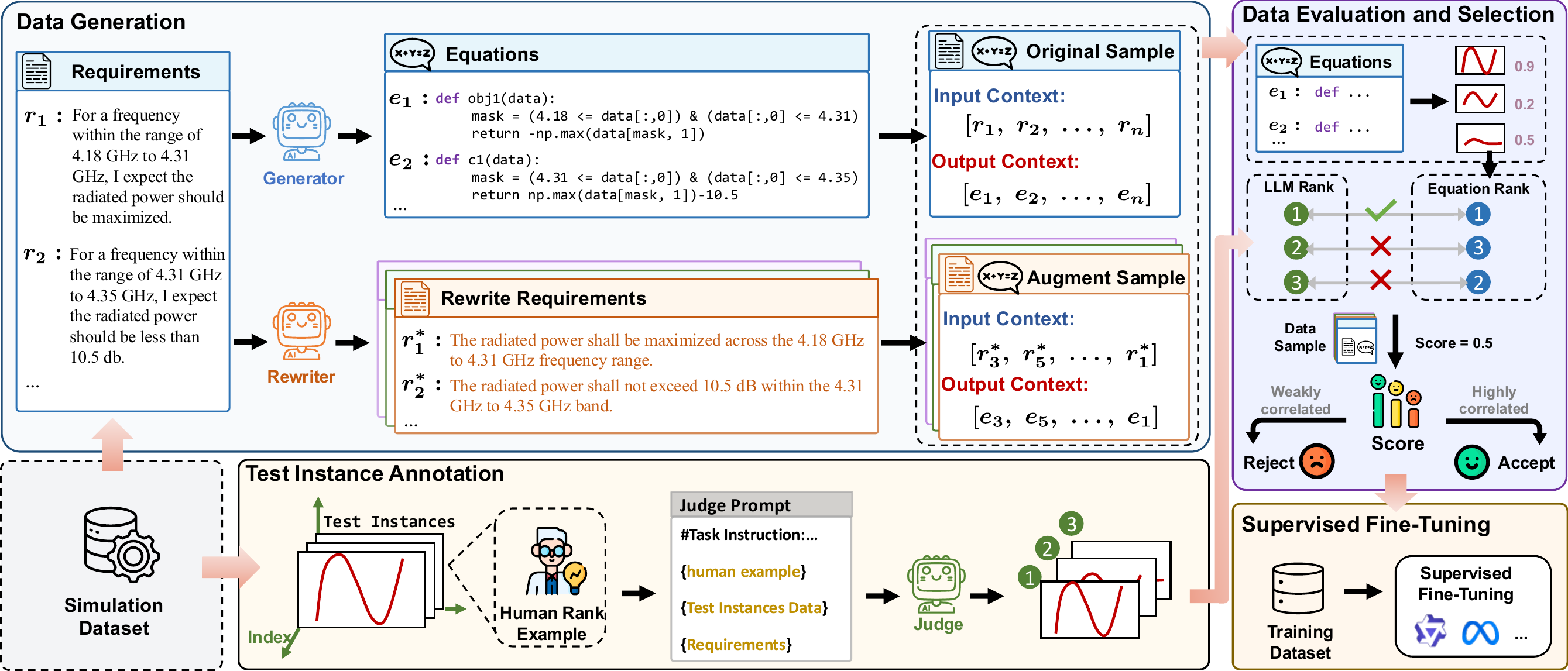}
    \caption{\textbf{Overview of the APF framework.} (a) \textit{Data Generation:} Design requirements are derived from the simulation dataset and rewritten by the LLM to produce corresponding model equations. (b) \textit{Test Instance Annotation:} For each requirement, a set of test instances is generated and annotated with reference rankings by the LLM. (c) \textit{Data Evaluation and Selection:} Generated equations are evaluated against LLM-based rankings, and high-quality samples are selected to construct the training set.
    (d) \textit{Supervised Fine-Tuning:}
      Dataset is used to fine-tune an open-source LLM, significantly enhancing its capability to generate accurate and executable design formulations.}
    \label{fig:framework}
\end{figure*}

\section{Related Work}
\subsection{LLMs for Automated Problem Formulation}
The NL4Opt competition~\cite{ramamonjisonNL4OptCompetitionFormulating2023,ramamonjisonAugmentingOperationsResearch2022} provides an early benchmark for translating natural-language descriptions into formal optimization representations.
Early studies focused on leveraging zero-shot or few-shot reasoning through prompt engineering to automate the modeling process.
For instance, Chain-of-Experts \cite{xiaoChainofExpertsWhenLLMs2023} employs a multi-agent framework to decompose complex reasoning tasks, while OptiMUS \cite{ahmaditeshniziOptiMUSScalableOptimization2024} designs domain-specific prompt templates tailored to linear and mixed-integer programming.
More recent efforts have shifted toward fine-tuning LLMs through supervised or reinforcement learning. 
ORLM \cite{huangORLMCustomizableFramework2025} and OPTMATH \cite{luOptMATHScalableBidirectional2025} construct large-scale synthetic datasets using semi-automated pipelines to enable SFT. 
LLMOPT \cite{jiangLLMOPTLearningDefine2024} combines multi-instruction learning with alignment techniques to improve the generalization of LLMs  in mathematical modeling tasks. 
SIRL \cite{chenSolverInformedRLGrounding2025} further introduces reinforcement learning with verifiable rewards to enhance the reliability of LLMs in optimization problem formulation.
However, these approaches are often constrained by the need for extensive expert annotation or instantaneous solver feedback, both of which are impractical in simulation-driven design due to the scarcity of domain experts and the high computational cost of evaluation.

\subsection{LLM-Based Synthetic Data Generation}
In recent years, LLM-based synthetic data generation has been widely adopted across a range of tasks \cite{longLLMsDrivenSyntheticData2024, liuBestPracticesLessons2024,gu2026paretohqd}.  
For domain-specific fine-tuning, generating high-quality and task-relevant synthetic data is crucial for enhancing model performance and generalization \cite{wangSelfInstructAligningLanguage2023,zelikmanSTaRBootstrappingReasoning2022}.
Current approaches to synthetic data generation using LLMs can be broadly categorized into two types: one generates structured task data from scratch by leveraging the internal knowledge of LLMs \cite{liSyntheticDataAlmost2024,xuMagpieAlignmentData2024}; the other uses seed inputs to guide the generation process, allowing for greater control and diversity in the output \cite{luoWizardCoderEmpoweringCode2023,mitraAgentInstructGenerativeTeaching2024,abdinPhi4TechnicalReport2024}.

In the field of automated theorem proving, the methods such as DeepSeek-Prover \cite{xinDeepSeekProverAdvancingTheorem2024} and LLM-ATPH \cite{laiLLMbasedAutomatedTheorem2025} construct formalized reasoning trajectories to produce high-quality synthetic datasets, significantly improving the proof capabilities of LLMs. 
In the domain of automated problem formulation, recent works like ORLM~\cite{huangORLMCustomizableFramework2025}, OptMATH~\cite{luOptMATHScalableBidirectional2025} and LLMOPT~\cite{jiangLLMOPTLearningDefine2024} have adapted these strategies to synthesize large-scale modeling datasets by combining expert labels with LLMs augmentation.
Nonetheless, their reliance on execution-based filtering is not scalable for simulation-driven design, where verification entails expensive physics simulations.

\section{The New Method}
\label{sec:method}

To address the misalignment between design intent and optimization models in simulation-driven design, we introduce APF. 
A primary challenge in this domain is the acquisition of reliable training data, where reliance on numerical solvers makes large-scale validation extremely expensive. 
APF addresses this with a test instance-based strategy that avoids solver-based verification and enables data selection without expensive simulations. 
We fine-tune LLMs to align natural language design specifications with formal modeling equations. 
As illustrated in Figure~\ref{fig:framework}, our framework consists of four integral modules: data generation, test instance annotation, data evaluation and selection, and supervised fine-tuning, culminating in a model capable of automating high-cost simulation problem formulation.

\subsection{Representation}
Both the data generation and test instance annotation modules rely on extracting design requirements and test instances from the historical simulation dataset. 
However, industrial specifications are typically unstructured and highly context-dependent.
To address this and ensure the framework's generalizability across various design contexts, we introduce a unified abstract definition for both requirements and instances in APF. 

\subsubsection{Natural Language Requirement Representation}
To standardize the design descriptions contained in the simulation dataset into consistent natural-language requirements, we formalize each requirement using a structured tuple:
\begin{equation}
    r = (\mathcal{Z}, M, \mathcal{C}),
\end{equation}
where $\mathcal{Z}$ is a specific subregion of the evaluation variable $z$ (e.g., the passband of a frequency domain), $M: z \in \mathcal{Z} \to \mathbb{R} $ is a metric function (e.g., radiation efficiency), and $\mathcal{C}$ specifies the design intent, such as a threshold constraint (e.g., $\min_{z\in \mathcal{Z}}M(z) \geq 1.5$) or an optimization goal (e.g., $\max_{z \in \mathcal{Z}} M(z)$).
While this formalization captures the essential meaning of each requirement, real-world engineering designs rarely hinge on just one. 
Instead, they often involve multiple requirements that may be interdependent or even conflicting. 
To handle this complexity, we define a complete design requirements set $\mathcal{R}$ composed of multiple requirement statements.
\begin{equation}
    \mathcal{R} = \{r_1, r_2, \dots, r_n\},
\end{equation}
To construct physically feasible requirement sets, we derive these values directly from historical simulation records specific to the design problem.
Instead of randomly synthesizing parameters, we treat existing simulation outcomes as design targets. 
Specifically, for a given simulation sample, we extract its operating conditions to define $\mathcal{Z}$ and its performance metrics to populate $\mathcal{C}$. 
This data-driven approach ensures that every constructed $\mathcal{R}$ corresponds to a valid, physically solvable design.

\subsubsection{Test Instance Representation}
We define a test instance $I$ as the performance response of a design solution derived from high-fidelity simulations. Formally, this is expressed as:
\begin{equation}
I =\mathcal{S}(\mathbf{x}),
\end{equation}
where $\mathbf{x}$ is design parameter, and $\mathcal{S}$ represents the computationally expensive solver (e.g., an electromagnetic full-wave simulator). 
Unlike simple scalar metrics, $I$ typically appears as a high-dimensional curve or vector over the evaluation domain. Since designers often impose specific constraints on different regions of this curve, the quality of a design depends on how well its response shape matches these targets. Given a requirement set $\mathcal{R}$, we can determine the quality of each instance based on its compliance. 
Thus, for a set of test instances $\mathcal{I} = \{I_1, I_2, \dots, I_m\}$, the requirements $\mathcal{R}$ determine a ranking $\pi_{\mathcal{R}}$, ordering the instances by their satisfaction of the design intent.

\subsection{Data Generation}
To effectively train LLMs for specialized domains like industrial design, the primary challenge lies in generating a large, diverse, and structured training dataset from ambiguous, human-expressed requirements. 
Based on the representation defined above, we construct a diverse set of training samples from the historical simulation dataset. Specifically, we first extract the structured requirement tuples from simulation records, and then leverage LLMs to generate the corresponding mathematical equations.
Moreover, we perform data augmentation to generate diverse training samples and enhance the generalization ability of LLMs.

\subsubsection{Equation Generation}
To translate natural language design requirements $\mathcal{R}$ into precise mathematical equations $E = \{e_1, e_2, \dots, e_n\}$, we proceed as follows. 
Our method leverages an LLM guided by a structured prompt template. 
This template embeds $\mathcal{R}$ along with clear instructions and rich contextual information, enabling the LLM to process the prompt and generate the corresponding mathematical formulation.
The detailed prompt template is provided in Appendix \ref{sec:data generation template}.
We generate a base dataset of $N$ samples, denoted as $D_{\text{base}} = \{(\mathcal{R}_i, E_i)\}_{i=1}^N$. 

\subsubsection{Data Augmentation}

We introduce a data augmentation strategy to align the dataset with simulation-driven design complexity. 
Our strategy couples semantic paraphrasing, which captures linguistic variations in requirements, with order permutation to model structural diversity inherent in engineering specifications.

\begin{enumerate}
\item \textbf{Semantic Paraphrasing.} 
To enhance robustness in handling diverse requirement descriptions, we employ an LLM to rewrite each requirement $r_i$ in the original set $\mathcal{R}$ into $v$ semantically equivalent variants. 
We constrain this process using prompt engineering to ensure that all variables, constants, and units remain unchanged. 
Subsequently, we randomly sample combinations of these variants to construct an augmented set $\mathcal{R}_{\text{aug}}$ containing $l$ design requirements.

\item \textbf{Order Permutation.} While the semantic meaning of requirements is invariant to their order, LLMs frequently demonstrate sensitivity to input sequences. 
To mitigate this, we randomly permute the requirements and their corresponding equations in tandem. 
This approach prevents the model from relying on spurious positional cues, ensuring it attends to the semantic content rather than the sequence order.

\end{enumerate}
Formally, for each base sample $(\mathcal{R}_i, E_i) \in D_{\text{base}}$, we generate $l$ augmented samples $\{(\mathcal{R}_{i,j}', E_i)\}_{j=1}^l$. 
The final training dataset is constructed by combining the base and augmented data: $\mathcal{D} = D_{\text{base}} \cup \bigcup_{i=1}^{N} \{(\mathcal{R}_{i,j}', E_i)\}_{j=1}^l$.

\subsection{Solver-independent Evaluation}
Given the modality gap between natural language requirements and mathematical equations, direct assessment of semantic alignment is computationally intractable.
To address this, we introduce a set of test instances $\mathcal{I}$ as a bridge, reframing the alignment challenge into a quantifiable ranking consistency problem.
Our core assumption is that an equation faithful to the design intent must yield an execution ranking on $\mathcal{I}$ that consistently matches the reference ranking defined by the requirements.

\subsubsection{Test Instance Annotation}

Obtaining a reliable reference ranking conditioned on specific requirements $\mathcal{R}$ is essential for evaluation.
Given the limited scalability of manual annotation, we use LLMs to automatically generate reference rankings.
Specifically, we adapt a listwise ranking strategy rather than pairwise comparisons.
Unlike pairwise methods that often lack global context, the listwise approach enables the LLM to evaluate the entire test instance set $\mathcal{I}$ jointly.
This perspective minimizes logical inconsistencies and captures complex trade-offs effectively.
We formulate this procedure as a conditional generation task guided by a structured prompt $\mathcal{P}$.
The detailed prompt template is provided in Appendix \ref{sec:test instance annotation template}.
To simulate expert reasoning, the prompt consists of four logical components:
\begin{equation}
    \mathcal{P} = \mathcal{P}_{\text{task}} \oplus \mathcal{P}_{\text{expert}} \oplus \psi_{\text{tab}}(\mathcal{I}) \oplus \psi_{\text{req}}(\mathcal{R}),
\end{equation}
where $\oplus$ denotes string concatenation, and $\psi(\cdot)$ represents the functions that convert structured data into text. 
The components are defined as follows:
\begin{enumerate}
\item Task Instruction $\mathcal{P}_{\text{task}}$: Provides the system-level prompt that defines the problem scope. 
It directs the LLM to function as a domain expert, outlining the logical procedure for balancing conflicting requirements to produce a reliable ranking.
\item Human Example $\mathcal{P}_{\text{expert}}$: Contains a one-shot human expert example. This part guides the model to focus on key features by showing how experts balance conflicting metrics.
\item Instances Data $\psi_{\text{tab}}(\mathcal{I})$: List the physical attributes of all test instances $\mathcal{I}$ in a table format. 
This structure supports horizontal comparison and helps generate the listwise ranking.
\item Requirements Query $\psi_{\text{req}}(\mathcal{R})$: Includes the current $\mathcal{R}$ in the context. This serves as the basis for ranking and ensures the evaluation standards fit the current task.
\end{enumerate}
Finally, the reference ranking $\pi_{\text{LLM}}$ is obtained by maximizing the posterior probability under the distribution $P_\theta$ of the LLM:
\begin{equation}\pi_{\text{LLM}} = \operatorname*{arg\max}_{\pi} P_\theta(\pi \mid \mathcal{P}).
\end{equation}
This generated sequence serves as the reference ranking for evaluating the quality of the generated equations.

\subsubsection{Data Evaluation and Selection}

To guarantee the reliability of our training data, we propose a ranking-based metric to evaluate the alignment between the generated formulation $E$ and the ground-truth design intent.
Given a set of test instances $\mathcal{I}$, we calculate the objective values and constraint violations using formulation $E$. The predicted ranking $\pi_{E}$ is derived through a hierarchical sorting strategy: feasible solutions are prioritized over infeasible ones, followed by non-dominated sorting based on Pareto dominance~\cite{yuConeConvexityCone1974}. 
This maps the mathematical properties of $E$ directly to a sequence of solution quality.
We define the quality score of $E$ as the Spearman correlation coefficient between the predicted ranking $\pi_{E}$ and the reference ranking $\pi_{\text{LLM}}$:
\begin{equation}
    S(E) = \rho(\pi_{E}, \pi_{\text{LLM}}),\label{eq:score}
\end{equation}
where a higher correlation implies that $E$ accurately captures the intent of the design requirements.

Based on this metric, we filter $\mathcal{D}_{\text{base}}$ to construct a high-quality dataset, $\mathcal{D}_{\text{HQ}}$. 
We only retain samples that show a strong correlation with the design requirements, discarding any pairs $(R_i, E_i)$ that fail to meet this standard. 
To ensure consistency, if a base sample is removed, its corresponding $l$ augmented samples are also excluded. 
Finally, $\mathcal{D}_{\text{HQ}}$ is used for model fine-tuning.

\section{Case Study: Antenna Design}

\begin{figure}[t]
    \centering
    \includegraphics[width=0.8\linewidth]{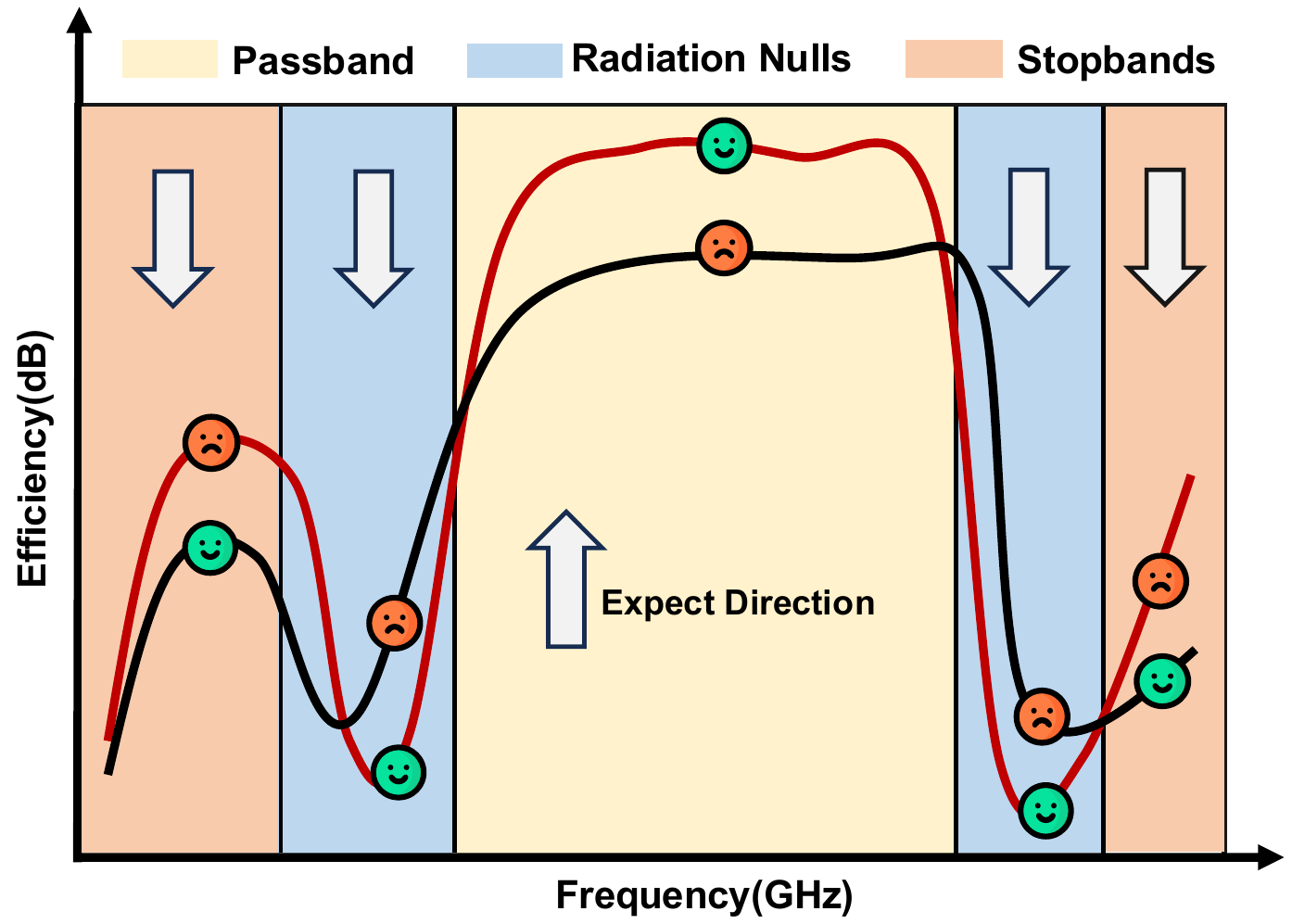}
    \caption{An example radiation efficiency curve is divided into five key frequency bands, each with distinct design requirements.}
    \label{fig:antenna_curve}
\end{figure}

To validate our method on a high-cost simulation-driven design task, we select antenna design as a case study. 
This problem involves optimizing structural parameters $x$ to shape the radiation efficiency curve across the frequency domain~\cite{elmisilmaniReviewDesignOptimization2020}. 
A critical challenge in this domain is the gap between high-level engineering intent and executable mathematical formulations. 
Engineers typically describe requirements qualitatively (e.g., demanding a flat passband or deep stopband rejection) rather than providing explicit objective functions.
Manually translating these ambiguous, multi-regional requirements into precise constraints for algorithmic optimization is both error-prone and labor-intensive.
Specifically, we focus on a representative three-layer filtering patch antenna~\cite{liangHighSelectivityHigh2022}. 
As illustrated in Figure \ref{fig:antenna_curve}, the radiation efficiency curve is characterized by five distinct frequency bands $\{\mathcal{Z}_{i}\}_{i=1}^5$: a Low Stopband, a Low Radiation Null, a Passband, a High Radiation Null, and a High Stopband.
This case constitutes a rigorous test for our method because the design goals across these bands are heterogeneous and strictly conflicting:
1) \textbf{Passband:} Maximize radiation efficiency while maintaining curve flatness to ensure stable signal transmission.
2) \textbf{Radiation Nulls:} Suppress efficiency to minimal levels within two narrow bands to filter specific interference.
3) \textbf{Stopbands:} Consistently suppress efficiency across wide flanking bands to prevent out-of-band noise.

\subsection{Experiment Settings}

To comprehensively assess APF, we conduct systematic comparisons against two mainstream categories of advanced methods. 
The first category includes prompting-based approaches, which evaluate the zero-shot reasoning ability of large language models. Specifically, we test direct zero-shot prompting on proprietary models (e.g., GPT-4o, DeepSeek-V3), as well as strong optimization-oriented prompting frameworks such as Chain-of-Experts \cite{xiaoChainofExpertsWhenLLMs2023} and OptiMUS \cite{ahmaditeshniziOptiMUSScalableOptimization2024}.
The second category covers our proposed fine-tuning-based APF methods. 
To ensure fairness, we adapt several state-of-the-art open-source LLMs as unified backbones, including Llama-3.1-8B-Instruct \cite{grattafioriLlama3Herd2024}, Qwen2.5-7B-Instruct \cite{qwenQwen25TechnicalReport2025}, and Mistral-7B-Instruct \cite{jiangMistral7B2023}. 
These models are used both for fine-tuning in APF and for evaluating prompting-based baselines, allowing us to directly assess the gains introduced by fine-tuning.
It is worth noting that other fine-tuning strategies such as ORLM \cite{huangORLMCustomizableFramework2025}, which require frequent interactions with external solvers, are excluded from our comparison due to the prohibitively high cost of running simulations in our industrial setting.

\begin{figure}[t]
    \centering
    \includegraphics[width=0.99\linewidth]{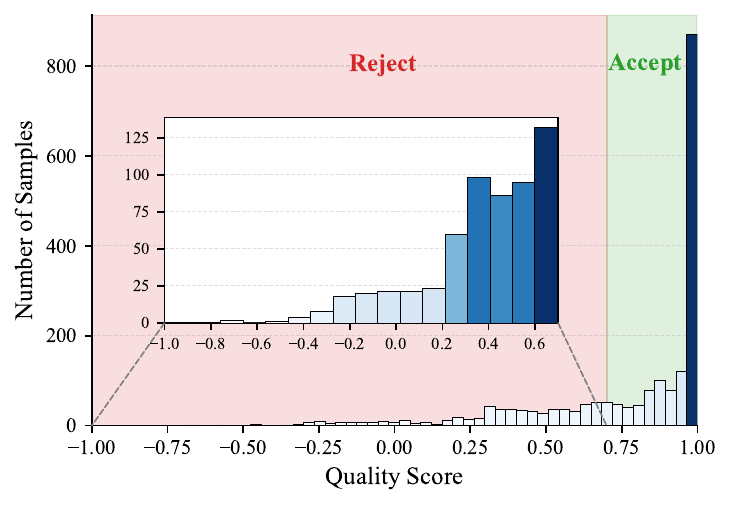}
    \caption{The distribution of quality scores for the samples.}
    \label{fig:distribution of data}
\end{figure}

We sampled 2,300 design requirement sets from historical antenna simulations. This collection was randomly split into a test set of 300 samples (ensuring zero overlap) and a candidate training set of 2,000 samples. Subsequently, APF generated corresponding mathematical equations specifically for the 2,000 training requirement sets to construct the base dataset $\mathcal{D}_{\text{base}}$.
To ensure the reliability of the training data, we analyzed the quality scores of $\mathcal{D}_{\text{base}}$, as shown in Figure \ref{fig:distribution of data}. The results indicate that most scores cluster between 0.9 and 1.0. This high consistency between the generated equations and design requirements confirms the effectiveness of our data generation process. 
To strictly maintain data quality for fine-tuning, we applied a selection threshold of 0.7, a value generally considered the lower bound for strong correlation. 
Samples exceeding this threshold were retained and subsequently augmented, resulting in a final high-quality dataset $\mathcal{D}_{\text{HQ}}$, comprising 7,879 samples.

\begin{figure*}[t] 
    \centering
    \includegraphics[width=0.95\linewidth]{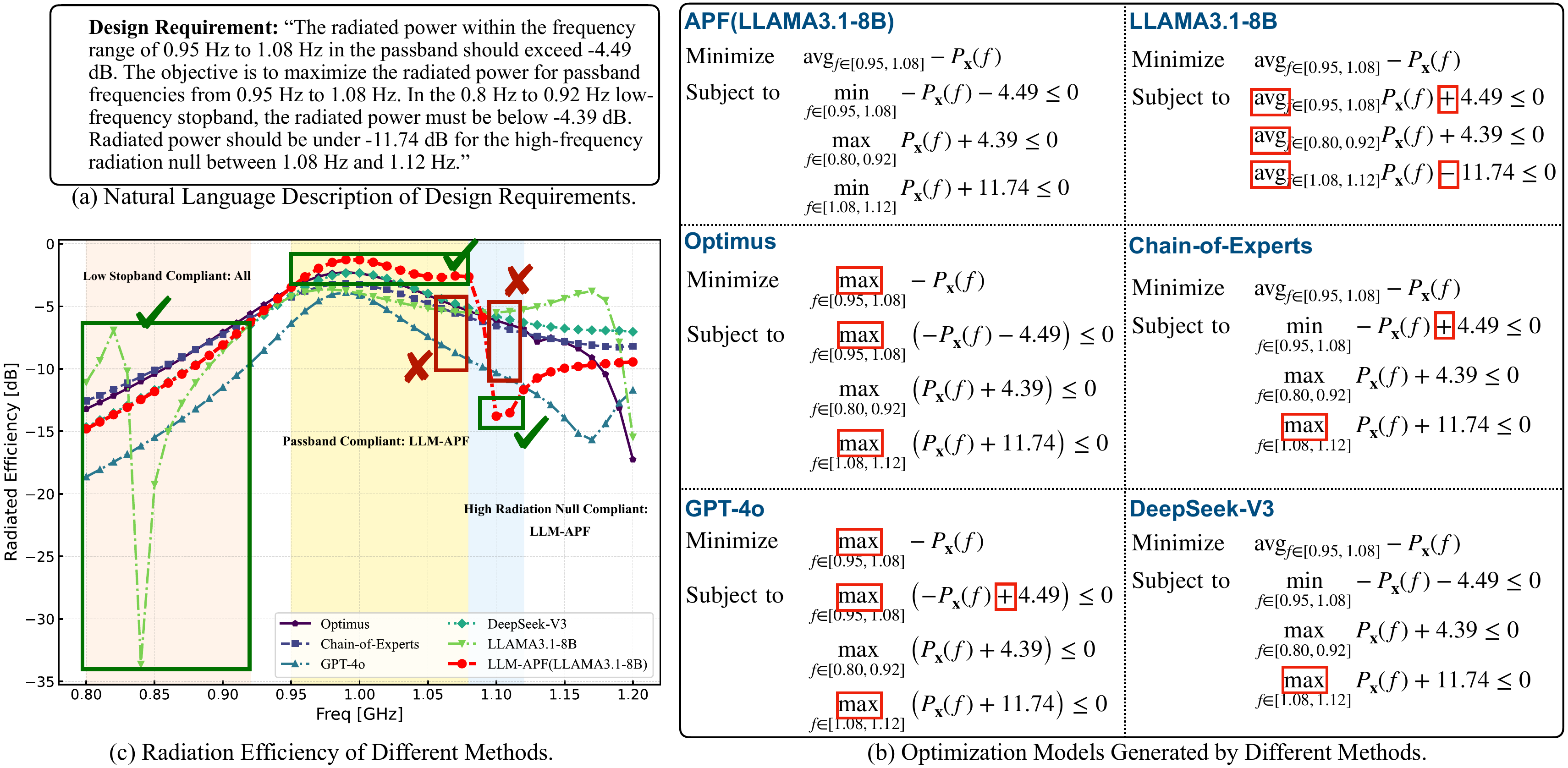}
    \caption{Comparison of radiation efficiency curves optimized using formulations generated by different methods.}
    \label{fig:method of antenna results}
\end{figure*}
\subsubsection{Evaluation Metrics}
We propose an alignment metric $A(E)$ to quantify how well the generated equation set $E$ satisfies the design requirements.
The set $E$ is partitioned into a subset of objective functions $E_{\text{obj}}$ (size $n_1$) and constraints $E_{\text{con}}$ (size $n_2$).
We conduct evaluations on a test instance set comprising $K$ selected antenna radiation efficiency curves.
For objective functions, the relative ordering of candidates is paramount.
We utilize the Spearman rank correlation ($\rho$) to measure the agreement between the ranking induced by each generated objective $e_i$ ($\hat{\pi}_i$) and the ground truth ranking ($\pi^*$):
\begin{equation}
    A_{\text{obj}}(E) = \frac{1}{n_1} \sum_{e_i \in E_{\text{obj}}} \rho(\hat{\pi}_i, \pi^*).
\end{equation}
For constraints, the focus is on the accurate determination of feasibility.
We define the alignment score as the classification accuracy. 
For each constraint $e_j$, we compute the agreement between the predicted feasibility vector $\hat{\mathbf{y}}_j$ and the ground truth $\mathbf{y}^*$:
\begin{equation}
    A_{\text{con}}(E) = \frac{1}{n_2} \sum_{e_j \in E_{\text{con}}} \left( 1 - \frac{1}{m} \| \hat{\mathbf{y}}_j - \mathbf{y}^* \|_1 \right),
\end{equation}
where $\| \cdot \|_1$ denotes the $L_1$ norm.
Finally, the total alignment score is calculated as $A(E) = \alpha A_{\text{obj}}(E) + (1 - \alpha) A_{\text{con}}(E)$, where the hyperparameter $\alpha \in [0, 1]$ balances the importance of objectives and constraints.
For all these metrics, a larger value indicates better performance. 
In our experiments, we set $\alpha = 0.5$.

\begin{table}[t]
\centering
\resizebox{\linewidth}{!}{
\begin{tabular}{lccc}
\toprule
\textbf{Method} & $\bm{A_{\textbf{obj}}}$ & $\bm{A_{\textbf{con}}}$ &  $\bm{A}$ \\
\midrule
\rowcolor{lightgray}
\multicolumn{4}{c}{\textit{Baselines}} \\ 
GPT-4o              & 0.6055 (0.2425)  & 0.7075 (0.2270)   & 0.6651 (0.1879) \\
DeepSeek-V3         & 0.7404 (0.2688)  & 0.7690 (0.1839)   & 0.7518 (0.2129) \\
Claude-sonnet-4.5 & \textbf{0.8023} (0.1128) & 0.7880 (0.1812) & 0.7923 (0.1346) \\
Chain-of-Experts    & 0.7426 (0.2410)  & 0.7453 (0.1865)   & 0.7252 (0.2309) \\
Optimus             & 0.6341 (0.2051)  & 0.6986 (0.2433)   & 0.6687 (0.1737) \\
\midrule
\rowcolor{lightgray}
\multicolumn{4}{c}{\textit{Open-Source LLMs}} \\
LLAMA3.1-8B         & -0.0453 (0.6470) & 0.5029 (0.1985)   & 0.2248 (0.4288) \\
Qwen2.5-7B          & 0.3542 (0.3208)  & 0.7333 (0.1129)   & 0.5292 (0.1458) \\
Mistral-7B          & 0.0733 (0.4403)  & 0.4936 (0.1713)   & 0.3007 (0.3644) \\
\midrule
\rowcolor{lightgray}
\multicolumn{4}{c}{\textit{Ours (APF) based on open-source LLMs}} \\
LLAMA3.1-8B & 0.8012 (0.1059)  & \textbf{0.7969} (0.1720)   & \textbf{0.7976} (0.1228) \\
Qwen2.5-7B  & 0.7990 (0.1106)  & 0.7959 (0.1739)   & 0.7961 (0.1243) \\
Mistral-7B  & 0.7974 (0.1171)  & 0.7883 (0.1912)   & 0.7918 (0.1345) \\
\bottomrule
\end{tabular}
}
\caption{Overall performance of APF fine-tuned models compared to baselines and open-source LLMs.
Results are reported as mean (std), with best in bold.}
\label{tab:model_comparison}
\end{table}

\subsection{Results}
\subsubsection{Quality of Generated Formulations}
We evaluate our proposed APF method by fine-tuning three open-source instruct models (LLAMA3.1-8B, Qwen2.5-7B, and Mistral-7B) on our dataset $\mathcal{D}_{\text{HQ}}$.
We then compare these fine-tuned models on the test set against four strong baselines and their original base models.
As shown in Table~\ref{tab:model_comparison}, our method consistently and significantly improves performance across all 7B and 8B models.
Significantly, the fine-tuned LLAMA3.1-8B model demonstrates the most notable improvement, with its overall score increasing from 0.2248 to 0.7976.
This represents a notable improvement over its base model and outperforms the evaluated baselines.

These results demonstrate two key findings. 
First, the high-quality data from APF significantly improve models' ability to formalize design requirements, overcoming limitations of pretrained models. 
Second, our approach enables smaller open-source models to match or exceed the performance of larger state-of-the-art models on domain-specific industrial design tasks.

\subsubsection{Antenna Design Performance}

To assess the real-world utility of our method, we design an optimization task based on the practical antenna design scenario.
The task requires LLMs to generate mathematical formulations for a constrained optimization problem, which aims to maximize passband radiation efficiency while adhering to strict power limits in stopband regions.
The natural language description of the design requirements is shown in Figure \ref{fig:method of antenna results}(a).
The mathematical formulations corresponding to the generated code of each method are illustrated in Figure \ref{fig:method of antenna results}(b).
We then feed each generated formulation into a scalable constrained Bayesian optimization (SCBO) solver \cite{erikssonScalableConstrainedBayesian2021} to obtain the final optimized antenna performance. The results are shown in Figure \ref{fig:method of antenna results}(c).
As shown in Figure \ref{fig:method of antenna results}(c), the radiation efficiency curve obtained from the optimization model generated by APF satisfies all design requirements. 
In contrast, the formulations from other methods fail to meet the design requirements for the passband and high radiation null. 
This demonstrates the superior ability of APF to accurately formulate design requirements into effective mathematical models, leading to optimized designs that better align with design intent.

\subsection{Robustness of Solver-Independent Evaluation}
\subsubsection{Judge Reliability}
To validate the reliability of the solver-independent evaluation, we performed a small-scale human ranking study. 
We sampled 30 design requirement sets and asked human evaluators to rank the same 15 test instances for each set. 
The rankings vary substantially across requirements despite sharing the same instance pool, indicating that the evaluation is genuinely requirement-conditioned.

We further compared human rankings with those produced by multiple LLM judges using Spearman's rank correlation. 
As shown in Table~\ref{tab:judge_agreement}, all judges achieve strong agreement with human preferences, including GPT-5 ($0.8316$), Gemini-3-Pro ($0.8175$), and GPT-5-mini ($0.8007$). 
These results indicate that the reference ranking is robust to the choice of judge model and that the LLM-as-a-judge protocol provides a reliable evaluation signal.

\begin{table}[t]
\centering
\small
\begin{tabular}{lcc}
\toprule
\textbf{LLM Judge} & \textbf{Mean } & \textbf{Std. } \\
\midrule
GPT-5          & \textbf{0.8316} & 0.1415 \\
Gemini-3-Pro   & 0.8175          & \textbf{0.1067} \\
GPT-5-mini     & 0.8007          & 0.1560 \\
\bottomrule
\end{tabular}
\caption{Agreement between human rankings and different LLM judges on 30 requirement sets, measured by Spearman's rank correlation ($\rho$).}
\label{tab:judge_agreement}
\end{table}

\subsubsection{Listwise Evaluation}

We further validate the choice of listwise evaluation by comparing it with a standard pairwise ranking baseline on the same requirement sets and the same 15 test instances. 
In our setting, the instance set is intentionally moderate and diagnostic rather than large-scale, so it can be evaluated in a single prompt without practical context-length issues.

As shown in Table~\ref{tab:listwise_pairwise}, listwise evaluation achieves comparable ranking quality to pairwise comparison, with a slightly higher correlation to the reference ranking (0.8643 vs.\ 0.8536). The main advantage of listwise evaluation is efficiency: pairwise comparison requires 105 LLM calls for 15 instances, whereas listwise evaluation requires only one. This reduces the evaluation time from 2544.8 seconds to 97.66 seconds and lowers the API cost from \$0.47 to \$0.02.

These results suggest that, in our setting where a moderate diagnostic set of carefully selected instances is sufficient to expose structural formulation errors, listwise evaluation is a more practical choice than pairwise comparison because it preserves ranking quality while being substantially faster and cheaper.
\begin{table}[t]
\centering
\small
\setlength{\tabcolsep}{4.5pt}
\begin{tabular}{lcccc}
\toprule
\textbf{Method} & \textbf{$\bm{\rho}$} & \textbf{Calls} & \textbf{Time (s)} & \textbf{Cost (\$)} \\
\midrule
Listwise (ours) & \textbf{0.8643} & 1   & \textbf{97.66}  & \textbf{0.02} \\
Pairwise        & 0.8536          & 105 & 2544.8          & 0.47          \\
\bottomrule
\end{tabular}
\caption{Comparison between listwise and pairwise evaluation on the same requirement set with 15 test instances. Spearman's rank correlation $\rho$ measures agreement with the reference ranking.}
\label{tab:listwise_pairwise}
\end{table}

\subsection{Ablation Study}

We ablate Data Augmentation and Data Selection by removing each component and fine-tuning the same backbone.
Table~\ref{tab:ablation_study} shows that both components contribute to $A_{\text{obj}}$, $A_{\text{con}}$, and overall $A$.

\textbf{Impact of Data Augmentation}
As shown in Table \ref{tab:ablation_study}, removing the data augmentation module (denoted as "w/o Augmentation") leads to a decrease in performance across all metrics. 
This indicates that data augmentation positively influences the model's performance. 
By generating diverse phrasings of design requirements, it improves the model's generalization ability to handle varied user inputs.

\textbf{Impact of Data Selection}
According to Table \ref{tab:ablation_study}, removing the data selection module (denoted as "w/o Selection") also results in performance degradation. 
This demonstrates that filtering synthetic data is a crucial step in the fine-tuning process. 
The synthetic dataset may contain low-quality or inaccurate design requirement-equation pairs, which introduce noise. 
By filtering out these inferior examples, the data selection module ensures that the model focuses on high-quality data, leading to better overall performance.

\begin{table}[htbp]
\centering
\small
\begin{tabular}{lccc}
\toprule
\textbf{Method} & $\bm{A_{\textbf{\text{obj}}}}$ & \bm{$A_{\textbf{\text{con}}}$} & \bm{$A$} \\
\midrule
w/o Augmentation      & 0.7656          & 0.7555          & 0.7553          \\
w/o Selection       & 0.7603          & 0.7800          & 0.7653          \\
\midrule
APF & \textbf{0.8009} & \textbf{0.7971} & \textbf{0.7976} \\
\bottomrule
\end{tabular}
\caption{
    Ablation results of APF with Augmentation and Selection removed.
}
\label{tab:ablation_study}
\end{table}

\subsection{Sensitivity Analysis}
\subsubsection{Threshold Sensitivity}

To evaluate the robustness of the data selection procedure, we vary the selection threshold used to construct the fine-tuning set from 0.6 to 0.8 and evaluate the resulting models under the same protocol. 
As shown in Table~\ref{tab:threshold_sensitivity}, the performance remains highly stable across all three settings, with only negligible changes in $A_{\text{obj}}$, $A_{\text{con}}$, and the overall alignment score $A$.

These results indicate that APF is not sensitive to the exact choice of threshold within a reasonable range. In particular, the consistent performance around the default threshold of 0.7 suggests that the effectiveness of data selection does not rely on narrow hyperparameter tuning, but on the ability of the solver-independent evaluation pipeline to identify high-quality synthetic samples.

\begin{table}[t]
\centering
\small
\resizebox{\linewidth}{!}{
\begin{tabular}{cccc}
\toprule
\textbf{Threshold} & $\bm{A_{\textbf{obj}}}$ & $\bm{A_{\textbf{con}}}$ & $\bm{A}$ \\
\midrule
0.6 & 0.7998 (0.1076) & 0.7965 (0.1721) & 0.7968 (0.1233) \\
0.7 & \textbf{0.8012} (0.1059) & \textbf{0.7969} (0.1720) & \textbf{0.7976} (0.1228) \\
0.8 & 0.7994 (0.1090) & 0.7967 (0.1720) & 0.7967 (0.1237) \\
\bottomrule
\end{tabular}}
\caption{Sensitivity of APF to the selection threshold used for constructing the fine-tuning dataset. Results are reported as mean (std).}
\label{tab:threshold_sensitivity}
\end{table}

\subsubsection{Instance Order Sensitivity}
To assess sensitivity to instance order, we randomly permuted the serialized order of the same 15 instances five times and recomputed the ranking correlation.
% The resulting correlations are highly stable (mean $\rho=0.8514$, std.\ $=0.0105$), suggesting that the judge's ranking is driven primarily by instance content rather than prompt position.
The resulting correlations are highly stable (mean $\rho = 0.8514$, std.\ $\rho = 0.0105$), suggesting that the judge's ranking is driven primarily by instance content rather than prompt position.

\section{Conclusions}

This paper addresses the challenge of translating ambiguous natural language requirements into precise mathematical optimization models. 
We present APF, a solver-independent framework that enhances formulation accuracy by synthesizing high-quality datasets for fine-tuning, eliminating the need for costly solver feedback. 
Experimental results on antenna design demonstrate that APF significantly outperforms baselines, yielding design outcomes that align more closely with engineering intent. 
This work provides a scalable and efficient path for automating domain-specific problem formalization.

\section*{Limitations}
While APF shows promise in automating problem formulation, two constraints warrant mention. First, our current evaluation focuses exclusively on antenna design. Although this domain effectively represents high-cost, simulation-driven tasks, we plan to validate the framework in broader physics-based engineering fields, such as aerodynamics and structural optimization, to rigorously assess its cross-domain generalizability. Second, the solver-independent evaluation relies on constructing prompts with detailed test instances. This approach is inherently bounded by the context window of current LLMs, potentially limiting the capability to process highly complex problem descriptions or large-scale validation datasets.

\bibstyle{acl_natbib}
\bibliography{main}

\newpage
\appendix
\section{Prompt Templates for APF Framework}
\label{sec:appendix_prompts}
This appendix presents the specific prompt templates used in our APF framework. 
To ensure reproducibility and transparency, we provide the exact instructions input to the LLMs. 
These prompts cover two distinct phases of our pipeline:
\begin{enumerate}
    \item \textbf{Data Generation Phase:} Used to generate training dataset mapping natural language requirements to executable code (Section~\ref{sec:data generation template}).
    \item \textbf{Evaluation Phase:} Used to establish reference rankings for test instances via an LLM-as-a-Judge approach (Section~\ref{sec:test instance annotation template}).
\end{enumerate}

\subsection{Data Generation}\label{sec:data generation template}
This section provides the detailed prompt templates used for data generation in our framework.
For data augmentation, we instruct the language model to rewrite each technical requirement into multiple alternative phrasings while strictly preserving the core technical meaning, all numerical values, and units. This ensures that the model is exposed to a wide range of natural language expressions without altering the underlying semantics.
For equation generation, we provide a prompt that guides the model to translate natural language requirements into executable Python functions suitable for numerical optimization. The prompt specifies the expected input and output formats, as well as the conventions for defining objective and constraint functions.
The full prompt templates for both stages are shown in Listings~\ref{lst:da_prompt} and~\ref{lst:eg_prompt}.

\begin{lstlisting}[style=mypromptstyle, caption={Prompt for Data Augmentation}, label={lst:da_prompt}]
You are an expert antenna systems engineer with a strong command of technical English. Your task is to act as a writing assistant.

You will be given a numbered list of technical design requirements. For each requirement in the list, you must generate exactly {num_versions} distinct, alternative phrasings.

Critical Rules:
Preserve Core Meaning: The fundamental physical and technical meaning of each requirement must be preserved EXACTLY. Keep All Numbers and Units: All numerical values (e.g., frequencies like 2.45 GHz, power levels like -10 dB, thresholds like -4.5) and their units (GHz, Hz, dB, etc.) MUST remain UNCHANGED. Maintain Conditions: The type of condition ('less than', 'greater than', 'maximized', 'minimized') must not be altered.

Format:
I will provide the requirements as a numbered list. Your response MUST be a single, valid JSON object. Do not add any introductory text or explanations outside of the JSON structure. The keys of the JSON object must be the number from the input list. The value for each key must be a JSON array of strings, where each string is one of the rewritten versions of the requirement.

Now, please generate {num_versions} versions for the following requirements:
{requirements_text}

\end{lstlisting}

\begin{lstlisting}[style=mypromptstyle, caption={Prompt for Equation Generation}, label={lst:eg_prompt}]
You are an expert in engineering optimization and scientific computing. Your task is to act as a bridge between natural language engineering specifications and executable Python code. Translate the following engineering design requirements into Python functions suitable for a numerical optimization library like NumPy. Each function should be self-contained and operate on a 2D NumPy array.

Following requirements:
{requirements_text}

You should adhere to the following rules
General Rules:
1.  Your output MUST be a valid JSON array `[...]`.
2.  Each object in the array corresponds to one numbered design requirement from the input.
3.  Do not output any text or code outside of the main JSON array.

JSON Object Schema:
Each object in the array must contain the following five keys:
1.  `"requirement_index"` (integer): The original number of the requirement (e.g., 1, 2).
2.  `"function_type"` (string): Either `"objective"` or `"constraint"`. Choose the most appropriate one.
3.  `"function_name"` (string): The name of the function (e.g., `"obj1"`, `"c2"`).
5.  `"code"` (string): The complete, self-contained Python function. This string MUST be properly escaped for JSON, especially newlines (`\\n`) and quotes (`\\"`).

Python Function Rules:
1.  Signature: Use type hints, e.g., `def obj1(data: np.ndarray) -> float:`.
2.  Input: The function must accept a single argument `data`, which is a 2D NumPy array.
3.  Objectives: Must be for minimization. To maximize a metric, minimize its negative.
4.  Constraints: Must be satisfied when the function's return value is `< 0`.
5.  Dependencies: Use `numpy` for all array operations. Assume `import numpy as np` is already executed.

\end{lstlisting}

\subsection{Test Instance Annotation}\label{sec:test instance annotation template}

This section provides the detailed prompt template used for test instance annotation in our solver-independent evaluation.
Given a requirement set $\mathcal{R}$ and a collection of test instances $\mathcal{I}$ (simulated performance curves), we instruct the LLMs to act as a domain expert and produce a listwise ranking over all curves.
The prompt explicitly separates objective and constraint requirements and enforces a two-stage procedure: (1) check feasibility against all constraints; (2) rank feasible curves by how well they satisfy the objectives, while ranking infeasible curves by the degree of constraint satisfaction.
This structured instruction reduces logical inconsistencies and yields a reliable reference ranking $\pi_{\text{LLM}}$ for downstream correlation-based data selection.
The full prompt template is shown in Listing~\ref{lst:rank_prompt}.

\begin{lstlisting}[style=mypromptstyle, caption={Prompt for Test Instance Annotation}, label={lst:rank_prompt}]
Task Instruction:
You are {{Expert_Role}}. 
Your task is to comprehensively evaluate and rank multiple curves considering both objective and constraint requirements.

[Optional: Few-Shot Example]
{{Example_Section}}

Now, please evaluate the following curves:

Design Requirements:
Objective Requirements (to be optimized):
{{List_of_Objectives}}

Constraint Requirements (must be satisfied):
{{List_of_Constraints}}

Evaluation Strategy:
Rule 1: Check Feasibility. 
Evaluate each curve against all Constraint Requirements, separate the curves into two groups: 1.Feasible curves: those that satisfy all Constraint Requirements. 2.Violating curves: those that fail to meet one or more Constraint Requirements.
The feasible curves will be ranked later according to 'Rule 2: Rank the Satisfying Curves'.
The violating curves will be ranked later according to 'Rule 3: Rank the Violating Curves'.

Rule 2: Rank Feasible Curves. 
Among the curves that satisfy all Constraint Requirements, rank them solely based on how well they collectively satisfy all Objective Requirements, without further considering the constraints.

Rule 3: Rank Violating Curves. 
If some curves violate the constraints, they should be ranked at the bottom.
If all curves fail to satisfy every constraint, you should still rank them. In this case, rank them based on how closely they satisfy all constraint requirements overall.
The curve that most nearly meets the full set of constraints should be ranked highest within this group.
Do not rank them according to the objective requirements in this situation.

Curve Data (JSON Format):
Each curve is represented as an object with: "curve": identifier, "data": list of [{{Axis_X_Name}}, {{Axis_Y_Name}}]. {{Data_Description}}
```json
{{JSON_Data_of_Curves}}

\end{lstlisting}

\section{Detailed Settings of APF}\label{sec:supplementary}
In the data augmentation phase of APF, for a set of design requirements $\mathcal{R}$, the number of rewritten versions of the design requirements is set to 3, and we select $l=5$ design requirement sets from all possible combinations as the augmented result of $\mathcal{R}$.

We implement all model training using the PyTorch framework, with base models being LLAMA3.1-8B, Qwen2.5-7B, and Mistral-7B, which have 7-8 billion parameters. We use 4 NVIDIA V100 Tensor Core GPUs with 32 GB each for model training and 4 NVIDIA V100 Tensor Core GPUs for model inference. The hyperparameters during training are shown in Table \ref{tab:sft_settings}.

\begin{table}[htbp]
    \centering
    \caption{Detail training settings for SFT.}
    \label{tab:sft_settings}
    \begin{tabular}{ll}
    \toprule
    \textbf{Parameter}      & \textbf{Value} \\
    \midrule
    LoRA\_Dropout           & 0.05           \\
    LoRA\_R                 & 16             \\
    LoRA\_Alpha             & 32             \\
    LearningRate            & 2.00E-04       \\
    BatchSize               & 16             \\
    MaxLength               & 1600           \\
    Epochs                  & 2              \\
    \bottomrule
    \end{tabular}
\end{table}

\section{Detailed Results}\label{sec:detailed_results}
In this section, we present the executable code generated by different automatic problem formulation methods for the same set of design requirements in a real antenna design task. 
We include results from several mainstream approaches, including APF (LLAMA3.1-8B), Optimus, Chain-of-Experts, GPT-4o, DeepSeek-V3, and the base LLAMA3.1-8B, covering both objective and constraint functions implemented in Python.

Each code example is given as a function, where the input is a two-dimensional NumPy array: the first column represents the evaluation variable (such as frequency), and the second column is the corresponding performance metric (such as radiation efficiency). The objective functions are implemented for minimization, and the constraint functions are satisfied when their return value is less than zero. These code samples allow for a direct comparison of the accuracy, readability, and practical usability of different methods in modeling.

\subsection{Results of APF(LLAMA3.1-8B)}
\begin{lstlisting}[style=mypython, caption={Code for APF(LLAMA3.1-8B)}]
def obj1(data: np.ndarray) -> float:
    """Minimizes the negative of the mean power in the 0.95-1.08 Hz passband."""
    mask = (data[:, 0] >= 0.95) & (data[:, 0] <= 1.08)
    if not np.any(mask):
        return 0.0
    passband_power = data[mask, 1]
    return -np.mean(passband_power)

def c1(data: np.ndarray) -> float:
    """Checks if the minimum power in the 0.95-1.08 Hz passband is greater than -4.49 dB."""
    mask = (data[:, 0] >= 0.95) & (data[:, 0] <= 1.08)
    if not np.any(mask):
        return 0.0
    min_power_in_passband = np.min(data[mask, 1])
    limit = -4.49
    return limit - min_power_in_passband

def c2(data: np.ndarray) -> float:
    """Checks if the maximum power in the 0.8-0.92 Hz stopband is less than -4.39 dB."""
    mask = (data[:, 0] >= 0.8) & (data[:, 0] <= 0.92)
    if not np.any(mask):
        return 0.0
    max_power_in_stopband = np.max(data[mask, 1])
    limit = -4.39
    return max_power_in_stopband - limit

def c3(data: np.ndarray) -> float:
    """Checks if the minimum power in the 1.08-1.12 Hz radiation zero points is less than -11.74 dB."""
    mask = (data[:, 0] >= 1.08) & (data[:, 0] <= 1.12)
    if not np.any(mask):
        return 0.0
    min_power_in_zero_points = np.min(data[mask, 1])
    limit = -11.74
    return min_power_in_zero_points - limit
    
\end{lstlisting}

\subsection{Results of Optimus}
\begin{lstlisting}[style=mypython, caption={Code for Optimus}]

def passband_power_maximization(data: np.ndarray) -> float:
    # Requirement 2: Maximize the radiated power for passband frequencies (0.95 Hz to 1.08 Hz)
    mask = (data[:, 0] >= 0.95) & (data[:, 0] <= 1.08)
    power_in_range = data[mask, 1]
    if len(power_in_range) == 0:
        return np.inf
    return -np.max(power_in_range)

def passband_power_constraint(data: np.ndarray) -> float:
    # Requirement 1: Ensure power > -4.49 dB for passband frequencies (0.95 Hz to 1.08 Hz)
    mask = (data[:, 0] >= 0.95) & (data[:, 0] <= 1.08)
    power_in_range = data[mask, 1]
    if len(power_in_range) == 0:
        return np.inf
    return np.max(-power_in_range - 4.49)

def low_stopband_power_constraint(data: np.ndarray) -> float:
    # Requirement 3: Ensure power < -4.39 dB for low-frequency stopband (0.8 Hz to 0.92 Hz)
    mask = (data[:, 0] >= 0.8) & (data[:, 0] <= 0.92)
    power_in_range = data[mask, 1]
    if len(power_in_range) == 0:
        return np.inf
    return np.max(power_in_range - (-4.39))

def high_radiation_zero_power_constraint(data: np.ndarray) -> float:
    # Requirement 4: Ensure power < -11.74 dB for high-frequency radiation zero points (1.08 Hz to 1.12 Hz)
    mask = (data[:, 0] >= 1.08) & (data[:, 0] <= 1.12)
    power_in_range = data[mask, 1]
    if len(power_in_range) == 0:
        return np.inf
    return np.max(power_in_range - (-11.74))
\end{lstlisting}

\subsection{Results of Chain-of-Experts}
\begin{lstlisting}[style=mypython, caption={Code for Chain-of-Experts}]

def obj2(data: np.ndarray) -> float:
    """Minimizes the negative of the mean power in the 0.95-1.08 Hz passband to maximize power."""
    mask = (data[:, 0] >= 0.95) & (data[:, 0] <= 1.08)
    passband_power = data[mask, 1]
    return -np.mean(passband_power)

def constraint1(data: np.ndarray) -> float:
    """Ensures the power is greater than -4.49 dB in the 0.95-1.08 Hz passband."""
    mask = (data[:, 0] >= 0.95) & (data[:, 0] <= 1.08)
    passband_power = data[mask, 1]
    return -np.min(passband_power) + 4.49

def constraint3(data: np.ndarray) -> float:
    """Ensures the power is less than -4.39 dB in the 0.8-0.92 Hz low-frequency stopband."""
    mask = (data[:, 0] >= 0.8) & (data[:, 0] <= 0.92)
    stopband_power = data[mask, 1]
    return np.max(stopband_power) + 4.39

def constraint4(data: np.ndarray) -> float:
    """Ensures the power is less than -11.74 dB in the 1.08-1.12 Hz high-frequency radiation zero points."""
    mask = (data[:, 0] >= 1.08) & (data[:, 0] <= 1.12)
    zero_points_power = data[mask, 1]
    return np.max(zero_points_power) + 11.74
\end{lstlisting}

\subsection{Results of GPT-4o}
\begin{lstlisting}[style=mypython, caption={Code for GPT-4o}]

def maximize_passband_power(data: np.ndarray):
    mask = (data[:, 0] >= 0.95) & (data[:, 0] <= 1.08)
    selected_power = data[mask, 1]
    return -np.max(selected_power)

def constraint_passband_power(data: np.ndarray):
    mask = (data[:, 0] >= 0.95) & (data[:, 0] <= 1.08)
    selected_power = data[mask, 1]
    return np.max(-selected_power + 4.49)

def constraint_low_frequency_stopband(data: np.ndarray):
    mask = (data[:, 0] >= 0.8) & (data[:, 0] <= 0.92)
    selected_power = data[mask, 1]
    return np.max(selected_power + 4.39)

def constraint_radiation_zero_points(data: np.ndarray):
    mask = (data[:, 0] >= 1.08) & (data[:, 0] <= 1.12)
    selected_power = data[mask, 1]
    return np.max(selected_power + 11.74)
\end{lstlisting}

\subsection{Results of DeepSeek-V3}
\begin{lstlisting}[style=mypython, caption={Code for DeepSeek-V3}]

def objective_maximize_passband_power(data: np.ndarray) -> float:
    mask = (data[:, 0] >= 0.95) & (data[:, 0] <= 1.08)
    passband_power = data[mask, 1]
    return -np.mean(passband_power)

def constraint_passband_min_power(data: np.ndarray) -> float:
    mask = (data[:, 0] >= 0.95) & (data[:, 0] <= 1.08)
    passband_power = data[mask, 1]
    return -np.min(passband_power) - 4.49

def constraint_low_stopband_max_power(data: np.ndarray) -> float:
    mask = (data[:, 0] >= 0.8) & (data[:, 0] <= 0.92)
    stopband_power = data[mask, 1]
    return np.max(stopband_power) + 4.39

def constraint_high_stopband_max_power(data: np.ndarray) -> float:
    mask = (data[:, 0] >= 1.08) & (data[:, 0] <= 1.12)
    stopband_power = data[mask, 1]
    return np.max(stopband_power) + 11.74
\end{lstlisting}

\subsection{Results of LLAMA3.1-8B}
\begin{lstlisting}[style=mypython, caption={Code for LLAMA3.1-8B}]


def obj1(data: np.ndarray) -> float:
    """Minimizes the negative of the mean power in the 0.95-1.08 GHz passband."""
    mask = (data[:, 0] >= 0.95) & (data[:, 0] <= 1.08)
    passband_power = data[mask, 1]
    return -np.mean(passband_power)

def con1(data: np.ndarray) -> float:
    """Ensures the power in the 0.95-1.08 GHz passband is greater than -4.49 dB."""
    mask = (data[:, 0] >= 0.95) & (data[:, 0] <= 1.08)
    passband_power = data[mask, 1]
    return np.mean(passband_power) + 4.49

def con2(data: np.ndarray) -> float:
    """Ensures the power in the 0.80-0.92 Hz stopband is less than -4.39 dB."""
    mask = (data[:, 0] >= 0.80) & (data[:, 0] <= 0.92)
    stopband_power = data[mask, 1]
    return np.mean(stopband_power) - 4.39

def con3(data: np.ndarray) -> float:
    """Ensures the power in the 1.08-1.12 Hz stopband is less than -11.74 dB."""
    mask = (data[:, 0] >= 1.08) & (data[:, 0] <= 1.12)
    stopband_power = data[mask, 1]
    return np.mean(stopband_power) - 11.74
\end{lstlisting}

\end{document}